\documentclass[10pt, a4paper]{article}
\usepackage{lrec2022} 
\usepackage{multibib}
\newcites{languageresource}{Language Resources}
\usepackage{graphicx}
\usepackage{tabularx}
\usepackage{soul}
\usepackage{titlesec}
\titleformat{\section}{\normalfont\large\bfseries\center}{\thesection.}{1em}{}
\titleformat{\subsection}{\normalfont\SmallTitleFont\bfseries\raggedright}{\thesubsection.}{1em}{}
\titleformat{\subsubsection}{\normalfont\normalsize\bfseries\raggedright}{\thesubsubsection.}{1em}{}
\renewcommand\thesection{\arabic{section}}
\renewcommand\thesubsection{\thesection.\arabic{subsection}}
\renewcommand\thesubsubsection{\thesubsection.\arabic{subsubsection}}

\usepackage{epstopdf}
\usepackage[utf8]{inputenc}

\usepackage{hyperref}
\usepackage{xstring}

\usepackage{color}

\usepackage{tabularx}
\usepackage{ltablex}
\usepackage{CJKutf8}
\usepackage{graphicx}
\usepackage{multirow}
\usepackage{microtype}
\usepackage{multicol}
\usepackage{footmisc}
\usepackage{ctable}
\usepackage{amsmath}
\usepackage{newtxtext,newtxmath}
\usepackage[labelfont=bf]{caption}
\usepackage{subcaption}
\usepackage{mathrsfs}
\usepackage{adjustbox}
\usepackage{bbding}
\usepackage{pifont}
\usepackage{soul}

\title{Automatic Speech Recognition Datasets in Cantonese: \\ A Survey and New Dataset}

\name{Tiezheng Yu$^\star$\thanks{$^\star$ These authors contributed equally.}, Rita Frieske$^\star$, Peng Xu$^\star$ $^\dagger$\thanks{$^\dagger$ The work was done when the author was studying in The Hong Kong University of Science and Technology.}, Samuel Cahyawijaya$^\star$, \\ \large{\textbf{Cheuk Tung Shadow Yiu, Holy Lovenia, Wenliang Dai, Elham J. Barezi}} \\ \large{\textbf{Qifeng Chen, Xiaojuan Ma, Bertram E. Shi, Pascale Fung}}}
\address{The Hong Kong University of Science and Technology \\
\texttt{\{tyuah, peng.xu, scahyawijaya\}@connect.ust.hk}\\
\texttt{rita.frieske@ust.hk}
}
\abstract{
Automatic speech recognition (ASR) on low resource languages improves the access of linguistic minorities to technological advantages provided by artificial intelligence (AI). In this paper, we address the problem of data scarcity for the Hong Kong Cantonese language by creating a new Cantonese dataset. Our dataset, \textbf{M}ulti-\textbf{D}omain \textbf{C}antonese \textbf{C}orpus (MDCC), consists of 73.6 hours of clean read speech paired with transcripts, collected from Cantonese audiobooks from Hong Kong. It comprises philosophy, politics, education, culture, lifestyle and family domains, covering a wide range of topics. We also review all existing Cantonese datasets and analyze them according to their speech type, data source, total size and availability. We further conduct experiments with Fairseq S2T Transformer, a state-of-the-art ASR model, on the biggest existing dataset, Common Voice zh-HK, and our proposed MDCC, and the results show the effectiveness of our dataset. In addition, we create a powerful and robust Cantonese ASR model by applying multi-dataset learning on MDCC and Common Voice zh-HK.
\\ \newline }

\begin{document}

\maketitleabstract

\section{Introduction}
Automatic speech recognition (ASR) systems take the audio as input and convert it into text \cite{malik2021automatic}. Due to the popularization of deep learning, ASR technology has grown rapidly and has led to a significant improvement in recognizing many languages. For instance, ASR systems in English \cite{zhang2020pushing,xu2021self,baevski2020wav2vec} have been able to achieve a below 2\% word error rate (WER) on the LibriSpeech \cite{panayotov2015librispeech} corpus. A similar trend is also observed in research on Chinese ASR \cite{li2019end,winata2020lrt,zhang2020pushing}, exemplified by the improvement in the ASR model performance on the Aishell-1 \cite{bu2017aishell} corpus from 18.7\% character error rate (CER) down to 6.84\% within just two years. However, many languages (e.g., Gujarati, Hindi, Bengali, Amharic and Cantonese), including those that feature code-switching, are still lacking resources, and the performance of ASR systems in these languages is unsatisfactory \cite{winata21_interspeech,khare2021low,lovenia2021ascend}. Therefore, many methods in ASR have also been introduced and have shown promising results \cite{wang2021improved,lin2020exploringdl,winata2020mtl,winata2020learningfA}. Many of these achievements are due to the utilization of the most recent deep neural network architectures and high-performance parallel computing graphics cards. However, as deep learning techniques require large amounts of training data, the creation of ASR datasets is essential for model performance. Moreover, creating a speech recognition corpus will accelerate the development of ASR systems in corresponding languages.

Although around 88.9\% of Hong Kong's population are native Cantonese speakers, the Cantonese language is still struggling with a shortage of resources for building ASR systems. We present the most important speech resources in Cantonese in Table \ref{tab:cantonese_datasets}, showing their speech type used for building the dataset, data source, total size and availability of the dataset. From the table, we can see that not all of them are suitable to build robust ASR systems. 

To fill the research gap, we introduce a multi-domain ASR read corpus called \textbf{M}ulti-\textbf{D}omain \textbf{C}antonese \textbf{C}orpus (MDCC) for ASR research in Cantonese. Our corpus consists of 73.6 hours of clean read speech collected from various Hong Kong Cantonese audiobook sources. It contains philosophy, politics, education, culture, lifestyle, and family domains, and covers a wider range of topics than most of the other corpora. In addition, we perform experiments using a state-of-the-art ASR framework, Fairseq S2T Transformer \cite{wang2020fairseq}, on the two of the largest available datasets (MDCC and Common Voice zh-HK). The model achieves 10.15\% CER on the test set of our corpus, which indicates the effectiveness of our dataset. We also use joint training to create a more powerful and robust model for Cantonese ASR \footnote{https://github.com/HLTCHKUST/cantonese-asr}.


\begin{table*}[t]
\centering
\begin{adjustbox}{width=\linewidth,totalheight={\textheight},keepaspectratio}
\begin{tabular}{lcccc}
\toprule
\textbf{Name}  & \textbf{Speech Type}  & \textbf{Data source}  & \textbf{Size [hours]}  & \textbf{Availability}  \\ \midrule
HKCAC \cite{leung2001hkcac}  & Spont.  & Phone-in programs  & 8.1  & Non-Public \\ 
HKCanCor \cite{luke2015hong} & Spont.  & Chat                          & 30.0 & Cvasi-Public     \\
HKCC \cite{chin2015linguistics} & Spont. & Movie  & 35.0  & Cvasi-Public \\
CantoMap \cite{winterstein2020cantomap} & Read & MapTask  & 12.8  & Public\\
Common Voice zh-HK \cite{ardila2019common} & Read  & Wikipedia  & 96.0  & Public\\
MDCC (Ours) & Read  & Audiobook & 73.6 & Public\\ \bottomrule
\end{tabular}
\end{adjustbox}
\caption{Hong Kong Cantonese ASR corpora}
\label{tab:cantonese_datasets}
\end{table*}

The contributions of our study to the field are threefold:
\begin{itemize}
    \item We review existing Cantonese ASR datasets and thoroughly analyze them from various perspectives (speech type, data source, total size and availability).
    \item We propose a new dataset named MDCC for ASR research in Cantonese, which consists of 73.6 hours of clean read speech and covers a wide range of topics.
    \item We evaluate our dataset and another available Cantonese ASR dataset (Common Voice zh-HK) by using a state-of-the-art ASR model (Fairseq S2T Transformer). Furthermore, we apply multi-dataset learning approaches on the two datasets to create a powerful and robust model for the Cantonese ASR. Multi-dataset learning boosts the model's performance on both datasets. The results support the effectiveness of our dataset.
\end{itemize}

\section{Cantonese ASR Datasets}
\label{sec:cantonese_asr_datasets}

Table \ref{tab:cantonese_datasets} lists the most important previous Cantonese ASR corpora with their speech type, data source, data size and availability. Some other works are also related to Cantonese ASR but focus on different aspects. For instances, Cantonese multimodal (audio-visual) speech dataset for in-car command recognition \cite{wenliang2022ci}.

\paragraph{HKCAC} The Hong Kong Cantonese Adult Language Corpus (HKCAC) is created from spontaneous speech records from the radio phone-in programs and forums in Hong Kong. It has 8.1 hours of recordings and transcripts that contain approximately 170,000 characters. The dataset is inaccessible as no website, link, or other information is provided for retrieving the dataset.

\paragraph{HKCanCor} The Hong Kong Cantonese Corpus (HKCanCor\footnote{http://compling.hss.ntu.edu.sg/hkcancor/}) is built based on spontaneous chat records. Participants were recruited for arranged recording sessions for two- or three-party chats. Later, an additional set of recordings was obtained from radio chat shows. The corpus consists of 30.0 hours of recordings, with each sample 10 minutes long. After transcription, the corpus contains around 180,000 word tokens. The transcripts of HKCanCor can be accessed from the official website, but no audio data are provided.

\paragraph{HKCC} The Corpus of Mid-20th Century Hong Kong Cantonese (HKCC\footnote{http://202.45.36.235/hkcc/}) is constructed based on Cantonese films from Hong Kong in the 1950s and 1960s. HKCC has two phases, and we only introduce the first-phase corpus since the second phase's report has not been released. There are 21 movies in the first-phase corpus, and each movie is about 100 minutes long. The corpus has, in total, about 200,000 character tokens. The details of the HKCC dataset can be found on the official website, but cannot be retrieved due to limited access control.

\paragraph{CantoMap} The Hong Kong Cantonese MapTask Corpus (CantoMap\footnote{https://github.com/gwinterstein/CantoMap}) aims to provide a Cantonese corpus for ASR research and also involves several controlled elicitation tasks related to the phonology and semantics of Cantonese. The design of the corpus follows the general setup used for the HCRC MapTask corpus \cite{anderson1991hcrc}. The corpus includes a total of 12.8 hours of recordings and transcripts of forty speakers. The CantoMap dataset is publicly available in the GitHub repository.

\paragraph{Common Voice zh-HK} The Common Voice zh-HK corpus\footnote{https://commonvoice.mozilla.org/zh-HK/datasets} is a massive-multilingual collection of transcribed speech collected and validated via Mozilla’s Common Voice initiative. The speakers are required to read sentences from Wikipedia and the annotators verify each sentence. We use  96.0 hour split of verified Cantonese utterances in our experiments. The detailed data statistics are shown in Section \ref{sec:experiments}. The dataset is available on the Common Voice website.

Although each of the existing corpora has advantages, not all of them are suitable for developing Cantonese ASR systems. None of the corpora except Common Voice zh-HK are large enough for data-intensive ASR model fine-tuning. Furthermore, even for Common Voice zh-HK, empirical experiments based on recent deep learning models are limited. To fill this research gap, we propose MDCC to enrich ASR data resources in Cantonese. Furthermore, we implement a state-of-the-art ASR model and report its performance on the Common Voice zh-HK  dataset and MDCC.

\section{Corpus Creation}
\label{sec:corpus_creation}
This section describes the creation of our MDCC. We first introduce our approach to collect and pre-process Cantonese audiobooks and then present the methods used to annotate transcripts.

\subsection{Audiobook Collection}
The speech corpus of the MDCC is collected from Hong Kong Cantonese audiobook sources. The corpus contains various audiobooks covering different topics (e.g., philosophy, politics, education, culture, lifestyle and family). Most of the books are in the literary form of Cantonese. However, some are written in a formal written form that is never used as spoken language, and therefore is not applicable to ASR systems. To remove these books from the dataset, we hire native Cantonese speakers to check all the audiobooks and filter them out manually.

Every chunk of the downloaded audiobook varies from 40 minutes to 2 hours, which does not fit the optimal size for ASR systems. Therefore, we apply a voice activity detection (VAD) tool to convert the original audio pieces into shorter audio utterances. The VAD tool can classify a chunk of audio data as being voiced or unvoiced, and we split the original audio samples at the unvoiced parts. After separation, we get 83,275 audio utterances with a total corpus size of 73.6 hours.

\subsection{Annotation}
\label{sec:annotation}
\begin{table}[t]
\centering
\begin{adjustbox}{width=\linewidth,totalheight={\textheight},keepaspectratio}
\begin{tabular}{p{0.31\linewidth} | p{0.69\linewidth}}
\toprule
\textbf{Type}        & \textbf{Words}  \\ \midrule
Unified Writing   & 
\begin{CJK*}{UTF8}{bsmi}
呢, 咧 => 呢 \hspace{1pt} - (Question particle) 
\newline呢邊, 呢便 => 呢邊 \hspace{1pt} - Over here / there;
\newline裏面, 裡面 => 裏面 \hspace{1pt} - Inside;
\end{CJK*} \\ \midrule 
Important Words & \begin{CJK*}{UTF8}{bsmi}
啦, 喇, 喎 \hspace{1pt} - (Question particle)
\newline係 \hspace{1pt} - Yes
\newline後 \hspace{1pt} - Later / After / Afterward
\newline幾 \hspace{1pt} - Some / A few / Several
\newline並 \hspace{1pt} - And
\newline併 \hspace{1pt} - Combine
\newline徵 \hspace{1pt} - Recruit / Ask for
\newline關於 \hspace{1pt} - About
\newline過程 \hspace{1pt} - Process
\newline盡快 \hspace{1pt} - As soon as possible
\newline部份 \hspace{1pt} - Partially
\newline咁樣 \hspace{1pt} - So / Such / Like this
\newline\includegraphics[width = 0.023\textwidth, height = 0.017\textheight]{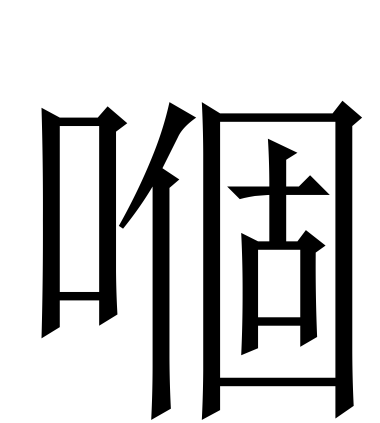}個 \hspace{1pt} - That one
\newline其實 \hspace{1pt} - Actually
\newline讀書 \hspace{1pt} - Study
\end{CJK*} \\ \bottomrule
\end{tabular}
\end{adjustbox}
\caption{The unified writing and important words for the attention of annotators.}
\label{tab:important_words}
\end{table}

To ensure cost-efficiency with optimal quality, we annotate all the utterances in two phases. We first conduct an automatic annotation with the Google Cloud Speech-to-Text API and then improve the quality of the automatic transcripts by hiring native Cantonese speakers to correct them manually.

\paragraph{Google Cloud Speech-to-Text.}
Google Cloud Speech-to-Text is an API that converts speech into text and is powered by Google’s AI technologies. \footnote{https://cloud.google.com/speech-to-text/} We apply this API to produce the initial transcripts of the utterances. More specifically, we use the default model with the language set to Hong Kong Cantonese (yue-Hant-HK). The API returns a transcript with a confidence score for each utterance. These automatically generated transcripts accelerate the hand-correction process significantly.

\paragraph{Proofreading of the Transcription.}
Since Google Cloud Speech-to-Text is not entirely accurate, we hire native Cantonese speakers to hand-correct the errors in the transcripts it generates. During proofreading, the annotators are required to adjust the transcripts and take notes for each utterance according to our guidelines. Table \ref{tab:important_words} gives the list of words that the annotators need to pay attention to. The words listed as ``unified writing" mean they share the same semantics, and we replace them with a unified word, while those listed as ``important words" need the special focus to annotate them accurately. One type of important words is question particles indicating interrogative sentences in Cantonese. Table \ref{tab:important_words} shows samples of question particles the annotators need to focus on.

\begin{table}[t]
\centering
\begin{adjustbox}{width=\linewidth,totalheight={\textheight},keepaspectratio}
\begin{tabular}{p{\linewidth}}
\toprule
\begin{CJK*}{UTF8}{bsmi}去其他地方啦\end{CJK*} - Go to other places\\ \midrule
\begin{CJK*}{UTF8}{bsmi}我有記憶嘅第一個冬天就係咁樣過去咗喇\end{CJK*} - The first winter that I can remember was gone\\ \midrule
\begin{CJK*}{UTF8}{bsmi}好多依然都係旗袍做校服\end{CJK*} - Many of them still wear cheongsams as school uniforms\\ \midrule
\begin{CJK*}{UTF8}{bsmi}由上海坐船到天津高級艙裏面亦都好多老鼠\end{CJK*} - There are also many mice in premium cabins of a ship going from Shanghai to Tianjin\\ \midrule
\begin{CJK*}{UTF8}{bsmi}我自己覺得好似跑得風一樣咁快\end{CJK*} - I feel like running as fast as the wind\\ \bottomrule
\end{tabular}
\end{adjustbox}
\caption{Several representative sentences from the MDCC.}
\label{tab:representative_sentences}
\end{table}

\begin{table*}[t]
\begin{adjustbox}{width=\linewidth,totalheight={\textheight},keepaspectratio}
\begin{minipage}[!b]{0.28\linewidth}
    \centering
    \includegraphics[width=\linewidth]{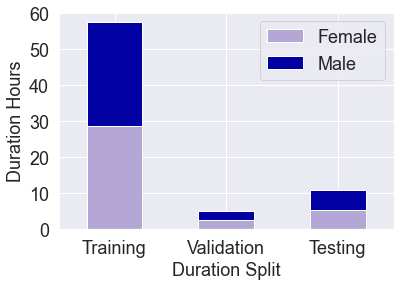}
    \captionof{figure}{Gender split of the training, validation and test sets per hour of recorded audio.}
    \label{fig:speaker-split}
\end{minipage}
\hfill
\hspace{6mm}
\begin{minipage}[b!]{0.65\linewidth}
\centering
\resizebox{\linewidth}{!}{
\begin{tabular}{lcccc|cccc} 
\toprule 
\multirow{2}{*}{Gender} & 
\multicolumn{4}{c}{\# Sample} & 
\multicolumn{4}{c}{Duration (hr)} \\
\cmidrule(l{2pt}r{2pt}){2-5} 
\cmidrule(l{2pt}r{2pt}){6-9}
&  Train & Valid  & Test  & Total & 
\multicolumn{1}{l}{Train} & 
\multicolumn{1}{l}{Valid.} & 
\multicolumn{1}{l}{Test} & Total \\ \midrule
Female & 29,224 & 2,541 & 5,606 & 37,371 & 28.67 & 2.52 & 5.39 & 36.58\\
Male & 35,896 & 3,122 & 6,886 & 45,904 & 28.86 & 2.54 & 5.61 & 37.01\\
Total & 65,120 & 5,663 & 12,492 & 83,275 & 57.53 & 5.05 & 11.01 & 73.59\\
\bottomrule
\end{tabular}}
\caption{Breakdown of the training, validation, and test splits in the MDCC by number of samples, gender, and the duration of the utterances.}
\label{tab:data_split}
\end{minipage}
\end{adjustbox}
\end{table*}

When taking notes, the annotators adhere to the following guidelines: 1) If the audio contains pure music, the annotators mark the label ``(music)" in the file name of its transcript. 2) If the utterance contains one or several sentences with background music or noise, the annotators mark the label ``(music)" before each sentence in the transcript. 3) The annotators use \{\} symbols to enclose words they are uncertain about, for example, \begin{CJK*}{UTF8}{bsmi}\{梁佳佳\}，我是\{\}人\end{CJK*}. 

In addition, for the English transcriptions or Arabic numerals, the annotators needed to do the following: 1) capitalize the first word of each complete English sentence; 2) capitalize proper nouns (e.g., names of people, countries and regions); 3) keep all other common English words lowercase except as set out in 1) and 2); 4) do not add a space between letters of common acronyms such as CCTV, VIP, etc, but do so between letters of unusual acronyms, and capitalize them, for example, \begin{CJK*}{UTF8}{bsmi}手機型號 X L 係幾多。\end{CJK*}; 5) add a space between English and Cantonese words; 6) convert all Arabic numbers to Cantonese based on pronunciation, for example, the Arabic number 60 can be converted into \begin{CJK*}{UTF8}{bsmi}六零 or 六十\end{CJK*} based on different pronunciations.

With the proofreading, the CER of the Google Cloud Speech-to-Text result reaches 25\%, proving that the annotators corrected numerous errors. Meanwhile, it is worthwhile to do annotation in two phases since most of the automatically generated transcripts from Google Cloud Speech-to-Text are in fact correct. In Table \ref{tab:representative_sentences}, we showcase several representative utterances. As we can see, most of the utterances contain a complete sentence, and the length of the utterances vary.

\subsection{Corpus Splitting}
\label{corpus_splitting}
We randomly split the MDCC into training, validation and test sets. Table \ref{tab:data_split} shows the detailed corpus splits which covers 65,120 utterances for training (57.53 hours), 5,663 for validation (5.05 hours), and 12,492 for testing (11.1 hours) respectively. As is shown in Figure \ref{fig:speaker-split}, we balance the total duration of each gender’s audio data within each split. 

\section{MDCC: Multi-Domain Cantonese Corpus}
\label{sec:multi-domain_cantonese_corpus}
In this section, we analyze our Multi-Domain Cantonese Corpus (MDCC) from the perspectives of data statistics, domain and text.

\subsection{Data Statistics}
\begin{figure}[t]
    \centering
    \includegraphics[width=\linewidth]{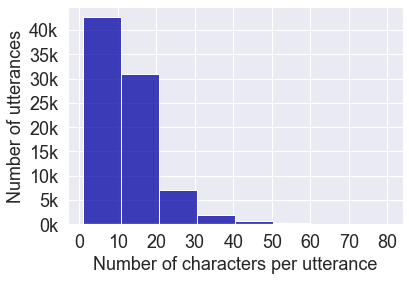}
    \caption{Distribution of the number of characters per utterance in the MDCC.}
    \label{fig:number_of_characters}
\end{figure}

The MDCC consists of 73.6 hours of Cantonese scripted speech from Cantonese native speakers, with a balanced gender ratio of 50.29\% male and 49.71\% female voice talents. The corpus is divided into 83,275 audio files, each containing one utterance. The MDCC includes a total of 998,366 Cantonese characters, with each utterance being approximately 11.99 characters long. As shown in Figure \ref{fig:number_of_characters}, the length of each utterance varies from a single character to as many as 80 characters. Of these utterances, 89.85\% are less than 23 characters, and the number of utterances decreases rapidly as the length of the utterance increases. Few utterances reach a length of more than 50 characters.

\begin{figure}[t]
    \centering
    \includegraphics[width=\linewidth]{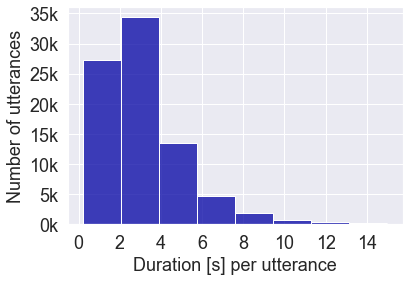}
    \caption{Distribution of the duration (in seconds) per utterance in our MDCC. }
    \label{fig:duration_length}
\end{figure}

The duration of each utterance is between 0.22 to 15.0 seconds. Moreover, the average duration of an utterance is 3.18 seconds. As we can see in Figure \ref{fig:duration_length}, the duration distribution is balanced, and most of the utterances are between one to nine seconds. Meanwhile, the duration distribution is generally aligned with the length distribution since longer utterances take more time for the speaker to read.

\begin{table*}[t]
\centering
\setlength\tabcolsep{10pt}
\begin{adjustbox}{width=\linewidth,totalheight={\textheight},keepaspectratio}
\begin{tabular}{c|lll}
\toprule
\textbf{Top}   & \textbf{Character/Unigram} & \textbf{Bigram} & \textbf{Trigram}\\ \midrule
1                    
& \includegraphics[width = 0.023\textwidth, height = 0.015\textheight]{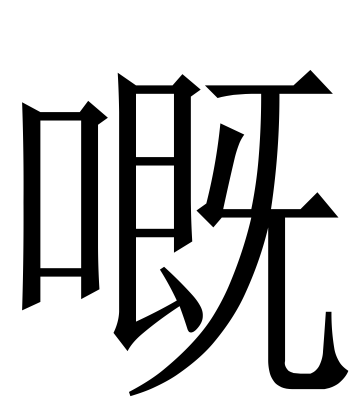} - is / are 
& \begin{CJK*}{UTF8}{bsmi}但係\end{CJK*} - but 
& \begin{CJK*}{UTF8}{bsmi}\includegraphics[width = 0.023\textwidth, height = 0.015\textheight]{figures/kai.png}時候\end{CJK*}  - the time of\\
2                    
& \begin{CJK*}{UTF8}{bsmi}一\end{CJK*}\hspace{1pt} - one                           
& \begin{CJK*}{UTF8}{bsmi}一個\end{CJK*} - one
& \begin{CJK*}{UTF8}{bsmi}呢一個\end{CJK*} - this one \\
3                    
& \begin{CJK*}{UTF8}{bsmi}係\end{CJK*}\hspace{1pt} - am / is / are                  
& \begin{CJK*}{UTF8}{bsmi}亦都\end{CJK*} - and / also / as well   
& \begin{CJK*}{UTF8}{bsmi}更懂得\end{CJK*} - more clear / understand better \\
4                    
& \begin{CJK*}{UTF8}{bsmi}佢\end{CJK*}\hspace{1pt} - he / she / it                  
& \begin{CJK*}{UTF8}{bsmi}同埋\end{CJK*} - with
& \begin{CJK*}{UTF8}{bsmi}同著作\end{CJK*} - same book\\
5                    
& \begin{CJK*}{UTF8}{bsmi}我\end{CJK*}\hspace{1pt} - I / me                         
& \begin{CJK*}{UTF8}{bsmi}佢\includegraphics[width = 0.021\textwidth, height = 0.015\textheight]{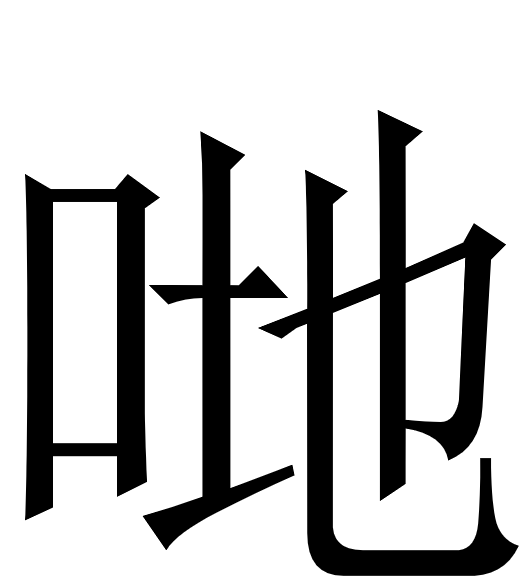}\end{CJK*} - they / them 
&  \begin{CJK*}{UTF8}{bsmi}自己\includegraphics[width = 0.023\textwidth, height = 0.015\textheight]{figures/kai.png}\end{CJK*} - myself / my own\\
6                    
& \begin{CJK*}{UTF8}{bsmi}有\end{CJK*}\hspace{1pt} - have / has                    
& \begin{CJK*}{UTF8}{bsmi}我\includegraphics[width = 0.023\textwidth, height = 0.015\textheight]{figures/die.png}\end{CJK*} - we / us
& \begin{CJK*}{UTF8}{bsmi}\includegraphics[width = 0.023\textwidth, height = 0.015\textheight]{figures/kai.png}學生\end{CJK*} - the student of\\
7                    
& \begin{CJK*}{UTF8}{bsmi}人\end{CJK*}\hspace{1pt} - person / people               
& \begin{CJK*}{UTF8}{bsmi}咁樣\end{CJK*} - like this 
& \begin{CJK*}{UTF8}{bsmi}唔能夠\end{CJK*} - cannot / unable to\\
8                    
& \begin{CJK*}{UTF8}{bsmi}唔\end{CJK*}\hspace{1pt} - no                            
& \begin{CJK*}{UTF8}{bsmi}就係\end{CJK*} - that is / just like 
& \begin{CJK*}{UTF8}{bsmi}中國人\end{CJK*} - Chinese  \\
9                    
& \begin{CJK*}{UTF8}{bsmi}個\end{CJK*}\hspace{1pt} - pieces                        
& \begin{CJK*}{UTF8}{bsmi}學生\end{CJK*} - student 
& \begin{CJK*}{UTF8}{bsmi}小王子\end{CJK*} - little prince    \\
10                   
& \begin{CJK*}{UTF8}{bsmi}好\end{CJK*}\hspace{1pt} - yes / good                    
& \begin{CJK*}{UTF8}{bsmi}裏面\end{CJK*} - inside 
& \begin{CJK*}{UTF8}{bsmi}呢一種\end{CJK*} - this kind  \\ \bottomrule
\end{tabular}
\end{adjustbox}
\caption{The statistics of the MDCC vocabulary: top 10 high-frequency unigrams, bigrams and trigrams.}
\label{tab:top10_characters}
\end{table*}

\subsection{Domain Analysis}
\label{subsec:domain_analysis}
\begin{figure}[t]
    \centering
    \includegraphics[width=\linewidth]{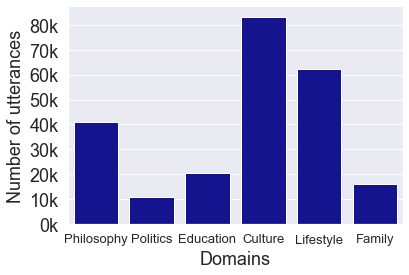}
    \caption{The distribution of domains of utterances in the MDCC. Each utterance can belong to more than one domain. Hence the total number of utterances per domain is bigger than the real total number per utterance. }
    \label{fig:domain_distribution}
\end{figure}

To create a general, natural and commonplace conversational ASR system, we choose a wide range of audiobook sources for the dataset. As a result, our MDCC dataset covers multiple domains, including philosophy, politics, education, culture, and lifestyle.  We hire a native Cantonese speaker to read and annotate the domain for each audiobook. Figure \ref{fig:domain_distribution} provides a summary of the domain distribution in the corpus. The domain of each sentence follows the domain of the audiobook that the sentence belongs to. Since an audiobook can cover multiple domains, the sum of the sentences in each domain is greater than the total number of sentences in the MDCC dataset. The culture and lifestyle domains have the most utterances in our dataset, which shows that the content of our dataset reflects people's daily lives. Besides culture and lifestyle, the philosophy domain also includes many utterances. It is worth explaining that ``philosophy" here mainly refers to self-help literature. The politics, education and family domains have less data, but are still considered essential topics of the MDCC. We believe that this domain analysis can help the research community to better understand the semantic distribution of our dataset.

\subsection{Common Phrases in MDCC}

\begin{figure}[t]
    \centering
    \includegraphics[width=\linewidth]{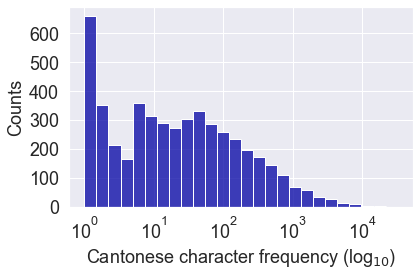}
    \caption{The distribution of log character frequency in the MDCC.}
    \label{fig:character_frequency}
\end{figure}


After a thorough analysis, we calculated a total of 998,366 Cantonese characters in the MDCC dataset. Their distribution is depicted in Figure \ref{fig:character_frequency}. In order to have an explicit understanding of the common phrases in the MDCC, we report the top 10 most common n-grams in Table \ref{tab:top10_characters}. A small proportion of the characters appear much more frequently than the others in the statistics. In detail, 14.56\% of the characters in the MDCC is made up of the  10 most common characters. Meanwhile, there is also a large number of characters that appear less than 10 times in the corpus, which reflects the diversity of the text in our dataset and its compliance with Zipf's law \cite{yu_zipf}.



\section{Cantonese ASR using Fairseq S2T Transformer}
\label{sec:experiments}
In this section, we introduce the Fairseq S2T Transformer \cite{wang2020fairseq,ott-etal-2019-fairseq} model and conduct experiments on the MDCC and Common Voice zh-HK dataset. We also apply multi-dataset learning approaches to further improve the model's performance.

\subsection{Fairseq S2T Transformer}
The reasons for choosing Fairseq S2T Transformer are twofold: 1) The model can achieve start-of-the-art performance on LibriSpeech, the de-facto standard ASR benchmark. 2) It is friendly for training custom models for ASR, and we can easily adapt the model to Cantonese. Based on the original transformer architecture \cite{vaswani2017attention}, Fairseq S2T Transformer proposes to add convolutional layers to the encoder \cite{mohamed2019}, which is the optimal way of processing audio data in the form of log mel filterbanks. We use the S2T Transformer XS version for all the experiments to implement the model.

\subsection{Datasets}

We conduct experiments on the two largest Cantonese datasets, MDCC and Common Voice zh-HK, for a better comparison, and jointly train them to see how the performance of the dataset improves if we double its size. The split of the Common Voice zh-HK dataset follows the same ratio as our MDCC, which is 80\% for training, 5\% for validation and 15\% for testing. The detailed information of these two datasets is shown in Table \ref{tab:common_vc_ours}. In addition, the audio files are downsampled to the frequency of 16kHz, with 32-bit depth. As follow-up experiments, we join the MDCC and Common Voice zh-HK datasets into one and apply a multi-dataset training approach. We hope that this can improve the model's performance in the cross-dataset setting and increase the robustness of the model.

\subsection{Implementation Details}
\paragraph{Data pre-processing.}  We implement spectral augmentation (SpecAugment), a state-of-the-art audio data augmentation method, which is implemented by masking certain frequency and time values on the spectrogram \cite{specaugment2019}. We use SpecAugment for the Common Voice zh-HK baseline, where it shows an improvement in overall results. Furthermore, we apply cepstral mean and variance normalisation (CMVN) for all the utterances \cite{strand2004cepstral}. In Fairseq S2T, pre-processed audio can be used directly or stored in the form of .npy files. The latter is the way in which we store features extracted from Cantonese datasets to achieve faster training. For tokenization of the transcribed data, we use the SentencePiece tokenizer \cite{kudo2018sentencepiece} with unigram subword tokenization \cite{kudo-2018-subword} and an 8,000-word vocabulary. The vocabulary covers 99.95\% of the characters in the MDCC (the default coverage for character-based languages).

\begin{table}[t]
\centering
\begin{adjustbox}{width=\linewidth,totalheight={\textheight},keepaspectratio}
\begin{tabular}{l|cccc}
\toprule
\textbf{Datasets}       & \textbf{\# train} & \textbf{\# val} & \textbf{\# test} &  \textbf{\# Total}\\ \midrule
Common Voice zh-HK                 & 65,437             & 4,089            & 12,269            & 81,795 \\
MDCC (ours) & 65,120             & 5,663            & 12,492            & 83.275 \\ \bottomrule
\end{tabular}
\end{adjustbox}
\caption{The statistics of the MDCC and Common Voice zh-HK, the two biggest Cantonese datasets, trained and tested with the Fairseq S2T Transformer model.}
\label{tab:common_vc_ours}
\end{table}

\paragraph{Hyper-parameters.}
    
We use the off-the-shelf Fairseq S2T Transformer XS model, which consists of a 6-layer encoder and a 3-layer decoder with a multi-head attention mechanism with four attention heads. For the objective function, we apply a cross-entropy loss with 0.1 label smoothing. The models are trained on four GPUs with a mini-batch size of 32. We use the default settings of SpecAugment provided in the S2T bundle: frequency masking width parameter F is set to 27, the number of time and frequency masks is set to 1,  with the upper bound of the time masking width 1; and the time masking width parameter T is set to 100. In our experiments, the applied SpecAugment policy does not include time warping.

\paragraph{Evaluation Metric.}
The models are evaluated on separated test sets, as is shown in Table \ref{tab:common_vc_ours}. For the evaluation, we average the performance of the last 10 checkpoints of the model using a beam search with beam size 8. Since the transcribed language is character based, we use the CER rather than WER as an evaluation metric \cite{Wang_CER}. The CER is calculated by adding the number of substituted, inserted and deleted characters together and dividing them by the total character count of the reference.

\subsection{Results and Analysis}
The CERs returned by S2T Transformer XS on the MDCC and Common Voice zh-HK datasets are comparable (10.15\% and 8.69\% CER respectively), possibly due to the similarities in domains and the size of the datasets themselves. We discover, however, that the datasets react differently to spectral augmentation. While SpecAugment hinder the training of the model on the MDCC, it does not influence the model trained on the Common Voice zh-HK dataset. Similarly the training on the joint dataset is hindered by adding SpecAugment. 
In the joint dataset we originally shuffled the training data, but the model did not converge. We conjecture that if we mix the datasets in the same batch, the model can not reach an optimal direction of gradient descent. The model benefits however from ordering the two datasets such that utterances from the MDCC are featured first and those from Common Voice zh-HK afterwards, supporting the theory that modelling long span word dependencies depends on the ordering of training data \cite{vazhenina2014sequence}. This decision is based on the fact that the MDCC data are cleaner, shorter and therefore easier to learn that those in Common Voice zh-HK. 

\begin{table}[t]
\begin{adjustbox}{width=\linewidth,totalheight={\textheight},keepaspectratio}
\begin{tabular}{l|ccc}
\toprule
       \textbf{Test set/Train set}           & \textbf{MDCC}  & \textbf{Common Voice zh-HK} & \textbf{Joint} \\ \midrule
\textbf{MDCC}         & 10.15 &  83.42   &    9.38 \\
\textbf{Common Voice zh-HK} & 53.44 &  8.69  &    7.65 \\ 
\textbf{Joint}     & 31.33 &  51.56  &    8.63 \\ \bottomrule
\end{tabular}
\end{adjustbox}
\caption{Character error rates (\%) returned by models trained on the MDCC, Common Voice zh-HK dataset and both combined datasets.}
\label{tab:baseline_results}
\end{table}

The CER returned by the model trained on the joint dataset shows improvement in the results of both datasets and a large improvement in the out-of-domain testing scenario. Even though the datasets used by us are the largest among all Cantonese ASR datasets, they are still much smaller compared with more comprehensive datasets such as LibriSpeech. Each of the Cantonese datasets contains fewer than 100 h of speech, while LibriSpeech alone contains 960 h of English audio \cite{panayotov2015librispeech}. Thus, combining both datasets is a natural step in creating strong Cantonese baselines for data-dependant deep learning models.


\section{Conclusion and Future Work}
\label{sec:conclusion_and_future_work}
In this paper, we review most of the previous Cantonese ASR corpora and thoughtfully analyze them. To address the limitations of the existing corpora, we propose a new dataset named the MDCC for the ASR research in the Cantonese language, which consists of 73.6 hours of clean read speech. We evaluate our dataset and compare it with the Common Voice zh-HK dataset using the Fairseq S2T Transformer model, and confirm that the results indicate the effectiveness of our proposed dataset. Our model trained on joint data outperforms Wav2Vec2-Large model on Cantonese dataset. \footnote{ \url{https://huggingface.co/ctl/wav2vec2-large-xlsr-cantonese} The data was not compared in the experiment section since the huggingface uses different splits of Common Voice zh-HK corpus.} For future work we plan to collect data from more audiobooks to enrich our dataset. In addition, we will create new Cantonese ASR corpora from different sources such as meetings and movies. Another future work direction is performing more experiments that combine the performance of the MDCC with multilingual datasets. We believe that our dataset and analysis can pave the way for future research works on the Cantonese ASR task and ASR for other low resource languages.

\section{Acknowledgements}
\label{sec:acknowledgements}
This  work  is  funded  by  ITS/353/19FP  of  the  Innovation Technology  Commission,  The  Hong  Kong  SAR  Government, School of Engineering Ph.D. Fellowship Award, The Hong Kong University of Science and Technology, and the Hong  Kong  Fellowship  Scheme  by  the  Hong  Kong  Research Grants Council (RGC).

\section{Bibliographical References}\label{reference}

\bibliographystyle{lrec2022-bib}
\bibliography{lrec2022-example}

\begin{thebibliography}{}

\bibitem[\protect\citename{Anderson \bgroup et al.\egroup
  }1991]{anderson1991hcrc}
Anderson, A.~H., Bader, M., Bard, E.~G., Boyle, E., Doherty, G., Garrod, S.,
  Isard, S., Kowtko, J., McAllister, J., Miller, J., et~al.
\newblock (1991).
\newblock The hcrc map task corpus.
\newblock {\em Language and speech}, 34(4):351--366.

\bibitem[\protect\citename{Ardila \bgroup et al.\egroup
  }2019]{ardila2019common}
Ardila, R., Branson, M., Davis, K., Henretty, M., Kohler, M., Meyer, J.,
  Morais, R., Saunders, L., Tyers, F.~M., and Weber, G.
\newblock (2019).
\newblock Common voice: A massively-multilingual speech corpus.
\newblock {\em arXiv preprint arXiv:1912.06670}.

\bibitem[\protect\citename{Baevski \bgroup et al.\egroup
  }2020]{baevski2020wav2vec}
Baevski, A., Zhou, H., Mohamed, A., and Auli, M.
\newblock (2020).
\newblock wav2vec 2.0: A framework for self-supervised learning of speech
  representations.
\newblock {\em arXiv preprint arXiv:2006.11477}.

\bibitem[\protect\citename{Bu \bgroup et al.\egroup }2017]{bu2017aishell}
Bu, H., Du, J., Na, X., Wu, B., and Zheng, H.
\newblock (2017).
\newblock Aishell-1: An open-source mandarin speech corpus and a speech
  recognition baseline.
\newblock In {\em 2017 20th Conference of the Oriental Chapter of the
  International Coordinating Committee on Speech Databases and Speech I/O
  Systems and Assessment (O-COCOSDA)}, pages 1--5. IEEE.

\bibitem[\protect\citename{Chin}2015]{chin2015linguistics}
Chin, A.
\newblock (2015).
\newblock A linguistics corpus of mid-20th century hong kong cantonese.
\newblock {\em Department of Linguistics and Modern Language Studies, The Hong
  Kong Institute of Education, Retrieved}, 23(3):2015.

\bibitem[\protect\citename{Dai \bgroup et al.\egroup }2022]{wenliang2022ci}
Dai, W., Cahyawijaya, S., Yu, T., Barezi, E.~J., Xu, P., Yiu, C. T.~S.,
  Frieske, R., Lovenia, H., Winata, G.~I., Chen, Q., Ma, X., Shi, B.~E., and
  Fung, P.
\newblock (2022).
\newblock Ci-avsr: A cantonese audio-visual speech dataset for in-car command
  recognition.

\bibitem[\protect\citename{Khare \bgroup et al.\egroup }2021]{khare2021low}
Khare, S., Mittal, A., Diwan, A., Sarawagi, S., Jyothi, P., and Bharadwaj, S.
\newblock (2021).
\newblock Low resource asr: The surprising effectiveness of high resource
  transliteration.
\newblock {\em Proc. Interspeech 2021}, pages 1529--1533.

\bibitem[\protect\citename{Kudo and Richardson}2018]{kudo2018sentencepiece}
Kudo, T. and Richardson, J.
\newblock (2018).
\newblock Sentencepiece: A simple and language independent subword tokenizer
  and detokenizer for neural text processing.
\newblock {\em arXiv preprint arXiv:1808.06226}.

\bibitem[\protect\citename{Kudo}2018]{kudo-2018-subword}
Kudo, T.
\newblock (2018).
\newblock Subword regularization: Improving neural network translation models
  with multiple subword candidates.
\newblock In {\em Proceedings of the 56th Annual Meeting of the Association for
  Computational Linguistics (Volume 1: Long Papers)}, pages 66--75, Melbourne,
  Australia, July. Association for Computational Linguistics.

\bibitem[\protect\citename{Leung and Law}2001]{leung2001hkcac}
Leung, M.-T. and Law, S.-P.
\newblock (2001).
\newblock Hkcac: the hong kong cantonese adult language corpus.
\newblock {\em International journal of corpus linguistics}, 6(2):305--325.

\bibitem[\protect\citename{Li \bgroup et al.\egroup }2019]{li2019end}
Li, M., Liu, M., and Masanori, H.
\newblock (2019).
\newblock End-to-end speech recognition with adaptive computation steps.
\newblock In {\em ICASSP 2019-2019 IEEE International Conference on Acoustics,
  Speech and Signal Processing (ICASSP)}, pages 6246--6250. IEEE.

\bibitem[\protect\citename{Lin and Chen}2020]{lin2020exploringdl}
Lin, W.-T. and Chen, B.
\newblock (2020).
\newblock Exploring disparate language model combination strategies for
  mandarin-english code-switching asr.
\newblock In {\em ROCLING}.

\bibitem[\protect\citename{Lovenia \bgroup et al.\egroup
  }2021]{lovenia2021ascend}
Lovenia, H., Cahyawijaya, S., Winata, G.~I., Xu, P., Yan, X., Liu, Z., Frieske,
  R., Yu, T., Dai, W., Barezi, E.~J., and Fung, P.
\newblock (2021).
\newblock Ascend: A spontaneous chinese-english dataset for code-switching in
  multi-turn conversation.

\bibitem[\protect\citename{Luke and Wong}2015]{luke2015hong}
Luke, K.~K. and Wong, M.~L.
\newblock (2015).
\newblock The hong kong cantonese corpus: design and uses.
\newblock {\em Journal of Chinese Linguistics}, 25(2015):309--330.

\bibitem[\protect\citename{Malik \bgroup et al.\egroup
  }2021]{malik2021automatic}
Malik, M., Malik, M.~K., Mehmood, K., and Makhdoom, I.
\newblock (2021).
\newblock Automatic speech recognition: a survey.
\newblock {\em Multimedia Tools and Applications}, 80(6):9411--9457.

\bibitem[\protect\citename{Mohamed \bgroup et al.\egroup }2019]{mohamed2019}
Mohamed, A., Okhonko, D., and Zettlemoyer, L.
\newblock (2019).
\newblock Transformers with convolutional context for {ASR}.
\newblock {\em CoRR}, abs/1904.11660.

\bibitem[\protect\citename{Ott \bgroup et al.\egroup
  }2019]{ott-etal-2019-fairseq}
Ott, M., Edunov, S., Baevski, A., Fan, A., Gross, S., Ng, N., Grangier, D., and
  Auli, M.
\newblock (2019).
\newblock fairseq: A fast, extensible toolkit for sequence modeling.
\newblock In {\em Proceedings of the 2019 Conference of the North {A}merican
  Chapter of the Association for Computational Linguistics (Demonstrations)},
  pages 48--53, Minneapolis, Minnesota, June. Association for Computational
  Linguistics.

\bibitem[\protect\citename{Panayotov \bgroup et al.\egroup
  }2015]{panayotov2015librispeech}
Panayotov, V., Chen, G., Povey, D., and Khudanpur, S.
\newblock (2015).
\newblock Librispeech: an asr corpus based on public domain audio books.
\newblock In {\em 2015 IEEE international conference on acoustics, speech and
  signal processing (ICASSP)}, pages 5206--5210. IEEE.

\bibitem[\protect\citename{Park \bgroup et al.\egroup }2019]{specaugment2019}
Park, D.~S., Chan, W., Zhang, Y., Chiu, C.-C., Zoph, B., Cubuk, E.~D., and Le,
  Q.~V.
\newblock (2019).
\newblock Specaugment: A simple data augmentation method for automatic speech
  recognition.
\newblock {\em Interspeech 2019}, Sep.

\bibitem[\protect\citename{Strand and Egeberg}2004]{strand2004cepstral}
Strand, O.~M. and Egeberg, A.
\newblock (2004).
\newblock Cepstral mean and variance normalization in the model domain.
\newblock In {\em COST278 and ISCA Tutorial and Research Workshop (ITRW) on
  Robustness Issues in Conversational Interaction}.

\bibitem[\protect\citename{Vaswani \bgroup et al.\egroup
  }2017]{vaswani2017attention}
Vaswani, A., Shazeer, N., Parmar, N., Uszkoreit, J., Jones, L., Gomez, A.~N.,
  Kaiser, {\L}., and Polosukhin, I.
\newblock (2017).
\newblock Attention is all you need.
\newblock In {\em Advances in neural information processing systems}, pages
  5998--6008.

\bibitem[\protect\citename{Vazhenina and Markov}2014]{vazhenina2014sequence}
Vazhenina, D. and Markov, K.
\newblock (2014).
\newblock Sequence memoizer based language model for russian speech
  recognition.
\newblock In {\em SLTU}, pages 183--187.

\bibitem[\protect\citename{Wang \bgroup et al.\egroup }2013]{Wang_CER}
Wang, P., Sun, R., Zhao, H., and Yu, K.
\newblock (2013).
\newblock A new word language model evaluation metric for character based
  languages.
\newblock In Maosong Sun, et~al., editors, {\em Chinese Computational
  Linguistics and Natural Language Processing Based on Naturally Annotated Big
  Data}, pages 315--324, Berlin, Heidelberg. Springer Berlin Heidelberg.

\bibitem[\protect\citename{Wang \bgroup et al.\egroup }2020]{wang2020fairseq}
Wang, C., Tang, Y., Ma, X., Wu, A., Okhonko, D., and Pino, J.
\newblock (2020).
\newblock Fairseq s2t: Fast speech-to-text modeling with fairseq.
\newblock {\em arXiv preprint arXiv:2010.05171}.

\bibitem[\protect\citename{Wang \bgroup et al.\egroup }2021]{wang2021improved}
Wang, D., Yu, J., Wu, X., Sun, L., Liu, X., and Meng, H.~M.
\newblock (2021).
\newblock Improved end-to-end dysarthric speech recognition via meta-learning
  based model re-initialization.
\newblock {\em 2021 12th International Symposium on Chinese Spoken Language
  Processing (ISCSLP)}, pages 1--5.

\bibitem[\protect\citename{Winata \bgroup et al.\egroup }2020a]{winata2020lrt}
Winata, G.~I., Cahyawijaya, S., Lin, Z., Liu, Z., and Fung, P.
\newblock (2020a).
\newblock Lightweight and efficient end-to-end speech recognition using
  low-rank transformer.
\newblock {\em ICASSP 2020 - 2020 IEEE International Conference on Acoustics,
  Speech and Signal Processing (ICASSP)}, pages 6144--6148.

\bibitem[\protect\citename{Winata \bgroup et al.\egroup }2020b]{winata2020mtl}
Winata, G.~I., Cahyawijaya, S., Lin, Z., Liu, Z., Xu, P., and Fung, P.
\newblock (2020b).
\newblock Meta-transfer learning for code-switched speech recognition.
\newblock In {\em Proceedings of the 58th Annual Meeting of the Association for
  Computational Linguistics}, pages 3770--3776, Online, July. Association for
  Computational Linguistics.

\bibitem[\protect\citename{Winata \bgroup et al.\egroup
  }2020c]{winata2020learningfA}
Winata, G.~I., Cahyawijaya, S., Liu, Z., Lin, Z., Madotto, A., Xu, P., and
  Fung, P.
\newblock (2020c).
\newblock Learning fast adaptation on cross-accented speech recognition.
\newblock In {\em INTERSPEECH}.

\bibitem[\protect\citename{Winata \bgroup et al.\egroup
  }2021]{winata21_interspeech}
Winata, G.~I., Wang, G., Xiong, C., and Hoi, S.
\newblock (2021).
\newblock {Adapt-and-Adjust: Overcoming the Long-Tail Problem of Multilingual
  Speech Recognition}.
\newblock In {\em Proc. Interspeech 2021}, pages 2451--2455.

\bibitem[\protect\citename{Winterstein \bgroup et al.\egroup
  }2020]{winterstein2020cantomap}
Winterstein, G., Tang, C., and Lai, R.
\newblock (2020).
\newblock Cantomap: a hong kong cantonese maptask corpus.
\newblock In {\em Proceedings of The 12th Language Resources and Evaluation
  Conference}, pages 2906--2913.

\bibitem[\protect\citename{Xu \bgroup et al.\egroup }2021]{xu2021self}
Xu, Q., Baevski, A., Likhomanenko, T., Tomasello, P., Conneau, A., Collobert,
  R., Synnaeve, G., and Auli, M.
\newblock (2021).
\newblock Self-training and pre-training are complementary for speech
  recognition.
\newblock In {\em ICASSP 2021-2021 IEEE International Conference on Acoustics,
  Speech and Signal Processing (ICASSP)}, pages 3030--3034. IEEE.

\bibitem[\protect\citename{Yu \bgroup et al.\egroup }2018]{yu_zipf}
Yu, S., Xu, C., and Liu, H.
\newblock (2018).
\newblock Zipf's law in 50 languages: its structural pattern, linguistic
  interpretation, and cognitive motivation.
\newblock {\em CoRR}, abs/1807.01855.

\bibitem[\protect\citename{Zhang \bgroup et al.\egroup }2020]{zhang2020pushing}
Zhang, Y., Qin, J., Park, D.~S., Han, W., Chiu, C.-C., Pang, R., Le, Q.~V., and
  Wu, Y.
\newblock (2020).
\newblock Pushing the limits of semi-supervised learning for automatic speech
  recognition.
\newblock {\em arXiv preprint arXiv:2010.10504}.

\end{thebibliography}


\end{document}